# Solving Minimum Vertex Cover Problem Using Learning Automata

Aylin Mousavian
Department of Electronic,
Computer and Biomedical
Engineering, Islamic Azad
University, Qazvin branch, Qazvin,
Iran
a.mousavian@qiau.ac.ir

Alireza Rezvanian
Computer Engineering &
Information Technology
Department, Amirkabir University
of Technology (Tehran
Polytechnic), Tehran, Iran
a.rezvanian@aut.ac.ir

Mohammad Reza Meybodi
Computer Engineering &
Information Technology
Department, Amirkabir University
of Technology (Tehran
Ploytechnic), Tehran, Iran
mmeybodi@aut.ac.ir

*Abstract*— **Minimum vertex cover problem is an NP-Hard problem with the aim of finding minimum number of vertices to cover graph. In this paper, a learning automaton based algorithm is proposed to find minimum vertex cover in graph. In the proposed algorithm, each vertex of graph is equipped with a learning automaton that has two actions in the candidate or non-candidate of the corresponding vertex cover set. Due to characteristics of learning automata, this algorithm significantly reduces the number of covering vertices of graph. The proposed algorithm based on learning automata iteratively minimize the candidate vertex cover through the update its action probability. As the proposed algorithm proceeds, a candidate solution nears to optimal solution of the minimum vertex cover problem. In order to evaluate the proposed algorithm, several experiments conducted on DIMACS dataset which compared to conventional methods. Experimental results show the major superiority of the proposed algorithm over the other methods.**

## I. INTRODUCTION

Set cover of undirected graph G=(V,E) is a subset S⊆V, such that every vertex v⊆V is either in S or adjacent to a vertex of S. In particular, vertex cover (VC) is set of vertices that cover all vertices in arbitrary graph, in other words, undirected graph G=(V,E), S⊆V is subset of vertices if (u,v)∈E then u∈S or v∈S. Vertex set with minimum cardinality is known as minimum vertex cover (MVC).
Minimum vertex cover problem (MVCP) is a NP-Complete problem [7, 8] which has exponential time complexity. Several practical applications of this problem include communication networks, social network [16], computer network and bioinformatics [1, 2]. MVCP introduced by Gary and Johnson and proved by Karp as NP-complete problem in connected graph in 1972. Therefore it is almost impossible finding polynomial time complexity algorithm for solving it. Thus many researchers have attracted to solve minimum vertex cover problem [9, 18]. Theoretical studied were carried out on the set cover by Cerderia and Pinto as focusing on the inequality conditions of vertices and edges in graphs [3, 19, 21].

In general, proposed solutions for MVCP are classified in two categories as heuristic and approximation algorithms [15]. Although, exact solution as branch and bound technique has been proposed for this problem, the time complexity is very high [4].
Heuristic algorithms for solving this problem have received many attractions for obtaining good approximate and near optimal results in reasonable time. For example, a greedy algorithm starts with an empty set, then, at each step, selects the best candidate with maximum degree and continues to reach the stop condition [3]. A popular approximation technique removes vertices of edges that have been chosen randomly from set. In this algorithm, the obtained result is always lower than twice the optimum solution, although this approximate method does not optimum solution but can be provided near optimal solution [5]. In [21], two approximation algorithms have been suggested for MVCP. The first method tries to select multiple nodes and uses minimum steiner tree to connect edges between them by minimum number of nodes to reach minimum vertex cover, thus another method must be applied to calculate minimum number of vertices of candidate set. This process increased running time. Second algorithm uses greedy technique similar to former method while the optimal number of nodes must be determined before as input of algorithm [21].
Another successful heuristic method for solving minimum vertex cover is genetic algorithm which finds a local optimum solution [4]. According to local optimality, the previous methods may provide better solution for MVC, but it is shown that genetic algorithm obtains answers in less time [4, 22]. Other successful method that can be referred here is parallel dynamic interaction (PDI), this method works much better than genetic and greedy methods in terms of convergence percentage of solution in consecutive iterations but has a higher running time and it does not provide a good result in large graphs [11].
In recent years, several learning automata based algorithms are presented to solve graph problems and successful results have been reported in literatures such as shortest path problem [10], minimum spanning tree problem [24], minimum dominating set problem [25], maximum clique problem [13, 6] and graph coloring problem [26].

In this paper, learning automata is applied to solve minimum vertex cover problem. In the proposed algorithm, a network of learning automata is created corresponding given graph, the collaboration between learning automata and learning mechanism that exist for each automaton, make the proposed algorithm able to obtain near optimal solutions.

The rest of paper is organized as follows. In section 2 learning automata and distributed learning automata are introduced briefly. In Section 3 proposed algorithm based on distributed learning automata is described in detail. The performance of proposed algorithm is evaluated through the simulation experiments in section 4. Finally section 5 concludes the paper.

## II. LEARNING AUTOMATA

Learning automaton (LA) [12, 14] is an abstract model of finite state machine so that it can perform finite actions which randomly chosen and evaluated by a stochastic environment and response is given as reward or penalty to LA. LA uses feedback of environment and select next action. During this process, LA learns how to chooses the best action from a set of allowed actions. Figure 1 illustrates relationship between the LA and the its random environment.

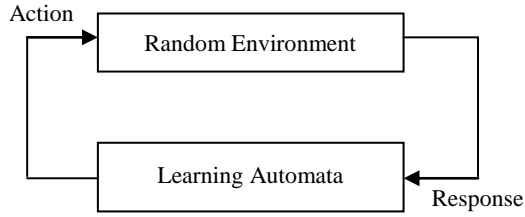

Fig. 1. The relationship between the LA and its random environment

LA with variable structure can be described with quadruple $\{\alpha, \beta, p, T\}$ where $\alpha=\{\alpha_1,\ldots,\alpha_r\}$ denotes the finite set of actions and $\beta=\{\beta_1,\ldots,\beta_m\}$ is the set of input values that can be taken by reinforcement signal and $p=\{p_1,\ldots,p_r\}$ is probability vector of each action. $P(n+1)=T[\alpha(n), \beta(n), p(n)]$ is learning algorithm where is updated in each iteration.

Equations (1) and (2) illustrate changes in probability vector according to performance evaluation of $\alpha_i$ in each step. Automaton updates its action probability set based on equation (1) for favorable responses as:

$$p_i(n+1) = p_i(n) + a[1-p_i(n)]$$
$$p_j(n+1) = (1-a)p_j(n) \qquad \forall j, \; j \neq i \quad (1)$$

And equation (2) for unfavorable ones:

$$p_i(n+1) = (1-b)p_i(n)$$
$$p_j(n+1) = \left(\frac{b}{r-1}\right) + (1-b)p_j(n) \qquad \forall j, \; j \neq i \quad (2)$$

Where $P(n)$ is the action probability vector at instant $n$. $r$ is the number of actions that can be taken by the LA. $a$ and $b$ denote reward and penalty parameters and determine the amount of increases and decreases of the action probabilities, respectively.

If a=b learning algorithm is called linear reward penalty ($L_{R-P}$), if $b<<a$ given learning algorithm is called linear reward-ε penalty ($L_{R-\varepsilon P}$), and finally if $b=0$ it is called linear reward inaction ($L_{R-I}$) [24-26].

### A. Distributed Learning Automata

A learning automaton is a simple agent for doing simple things. The full potential of learning automata is realized when they are interconnected to work together. A distributed learning automata (DLA) is a network of learning automata cooperating to solve a particular problem. In this network, each time just one automata is active. Number of actions performed by an automaton is equal to number of learning automata connected to it. By choosing an action of automata in this network, the connected automata at other side will be activated. The model of DLA network denotes a graph in each of vertex is an automaton [10].

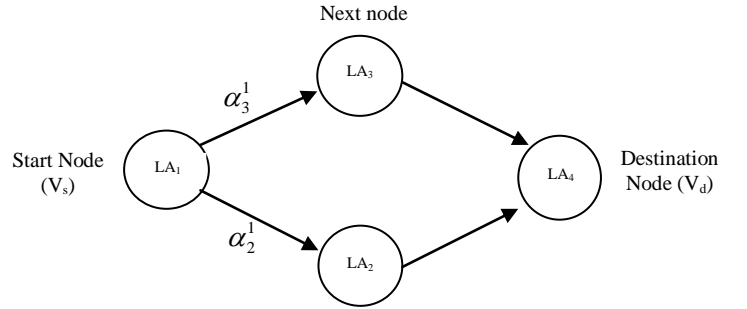

Fig. 2. Components of distributed learning automata

The edges (LA$_i$, LA$_j$) in graph means choosing action $\alpha_j^i$ cause activation of LA$_j$ by LA$_i$ [10, 13].

## III. THE PROPOSED ALGORITHM

In this section, a proposed algorithm based on learning automaton for solving minimum vertex cover problem is presented. Undirected graph G=(V,E) is served as input of algorithm and output is minimum expected vertex cover set. At first, a network of learning automata is formed by assigning an automaton to each vertex. Set of action that can be taken by learning automata is edges connected to each vertex of graph. This means the number of each automaton's action equal to number of set of edges connected to it.

Initially, it is assumed that all automatons are disabled. In first iteration, one automata is chosen randomly and activated, then chosen automata is added to covering set, All its neighbors are added to the set of neighborhood then this automata chooses one of its action (edges connected to automata). Selection of automata in first iteration is randomly because the probabilities of all edges are equal. This process is continued until the cardinality of covering set is equal to the predefined threshold. Chosen automata of each iteration placed in a separate category as path, then activated automata are known respectively.

After each iteration, number of vertices for covering set is compared with previous covering set vertices and automata with minimum set of vertices covering having maximum neighbors get rewarded, otherwise penalized them. Updating vector probability and learning mechanism function is mentioned in section 2. Stopping condition can be applied as a certain number of iterations to proposed algorithm. In this paper, in addition to specified number of iterations, the entropy of probability action vector is used as a criterion for stopping condition.

It is noticed that, at first, learning automata chooses its action with equal probability and will converge to an optimal value in learning process; this can be easily demonstrated by using entropy of action probability vector. Entropy is a fundamental concept in thermodynamic representing the degree of order/disorder in thermodynamic system that plays an important role in various fields of computer science, such as learning. Entropy in its basic indicates a measure of uncertainty action of learning automata at next selection. Higher entropy indicates greater degree of uncertainty at choosing next action. High uncertainty in the probability vector for learning automata means that automata have no useful information for achieving goal and select their actions randomly, but lower uncertainty measure indicates automata with higher likelihood select one of actions and have useful information to get the target. Suppose that $\{p_1, p_2, \ldots, p_r\}$ is the action probability vector of learning automata with r actions, entropy of action probability vector is determined as follows:

$$E(P) = -\sum_{i=1}^{r} p_i \log(p_i) \quad (3)$$

Entropy value will be highest when all actions probability is the same. At successive iterations entropy value will be came down [23]. In case of two-actions learning automata, entropy value will be between 0 and 1. For other case if number of actions is greater than two, it can be re-scaled between 0 to 1. The general structure of proposed algorithm is shown in the following diagram as figure 3.

Thus LA learns how to choose best action with minimum vertex cover in successive iterations. This process will be continued until the solution reach to near optimal. In next section, experimental simulations are demonstrated to evaluate proposed algorithm.

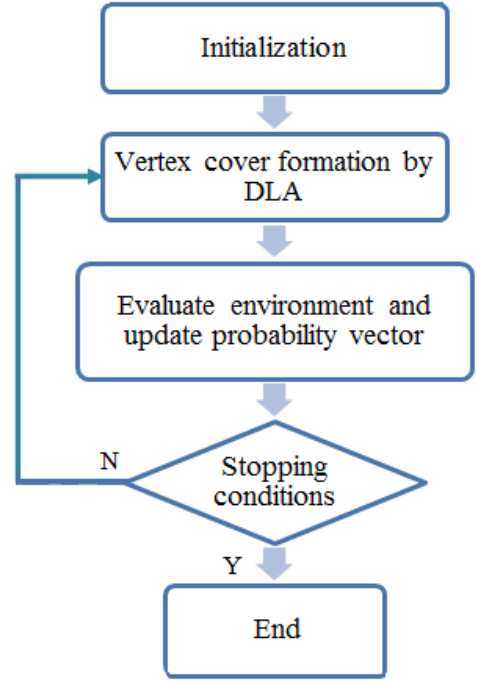

Fig. 3. Structure of Proposed Algorithm base on Learning Automata for Solving Minimum Vertex Cover

IV. SIMULATION RESULTS

To evaluate the proposed algorithm, DIMACS benchmark is used in experiments. For proposed algorithm, all experiments have been implemented using $L_{R-I}$ with learning rate 0.3 in learning algorithms. Results of proposed algorithm (VCLA) are compared with results of state of the art techniques such as EWLS [20] and EWCC [9] in table 1. The number of vertex cover in graphs is listed in table1. "Cn" indicates the number of covering nodes, "Lp" means the number of iterations required to achieve an acceptable answer. It is noted that successful rate of all algorithms for table 1 is obtained 100%

TABLE I. COMPARISON OF COVERAGE AND LOOPS FOR PROPOSED ALGORITHM WITH OTHER ALGORITHMS

| Algorithm | | CVLA | | EWCC | | EWLS | |
|---|---|---|---|---|---|---|---|
| Instances | # | Cn | Lp | Cn | Lp | Cn | Lp |
| Brock200 | 200 | 6 | 100 | 12 | 11780 | 12 | 11947 |
| C250 | 250 | 2 | 100 | 44 | 1743 | 44 | 1541 |
| C500 | 500 | 2 | 100 | 57 | 99203 | 57 | 11598 |
| Dsjc500 | 500 | 6 | 100 | 13 | 2344 | 13 | 3179 |
| Gen200p0944 | 200 | 2 | 100 | 44 | 1296 | 44 | 2434 |
| Gen200p0955 | 200 | 2 | 100 | 55 | 242 | 55 | 299 |
| Gen400p0955 | 400 | 2 | 100 | 55 | 29450 | 55 | 41906 |
| P_hat7002 | 700 | 13 | 119 | 44 | 212 | 44 | 222 |

According to results in the table 1, proposed algorithm (CVLA) has considerable success in terms of cover size and coverage loop.

Convergence of proposed method based on entropy changes on various benchmark graphs as is shown in figure 4. The rate of entropy changes were tested an average of 10 independent run for 1000 iteration.

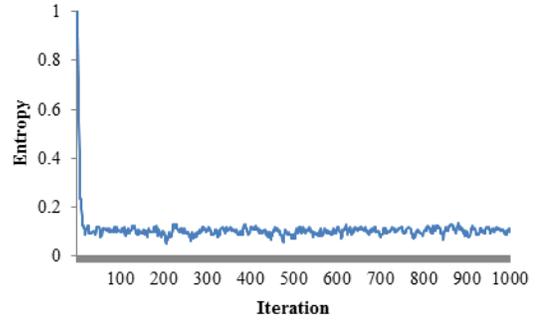

d) Brock200

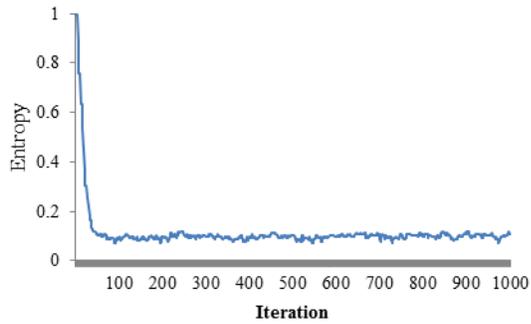

a) P_hat7002

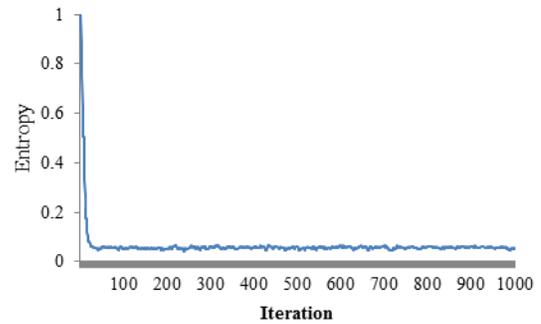

e) C250

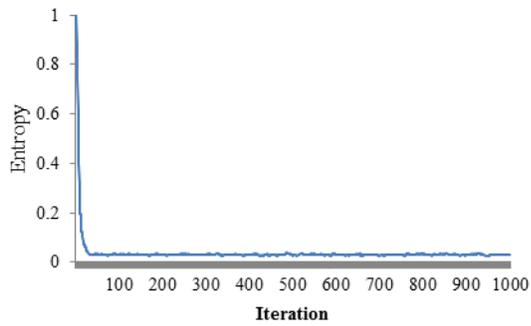

b) C500

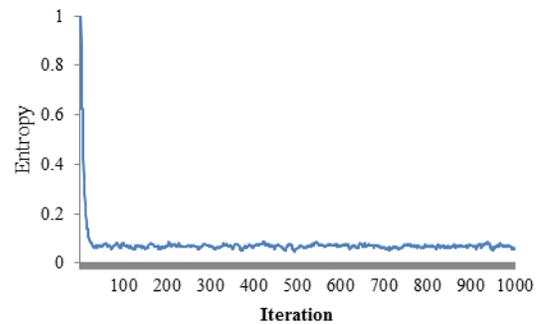

f) Gen200p0955

Fig. 4. Entropy of action probability vector of LA belong to minimum vertex cover

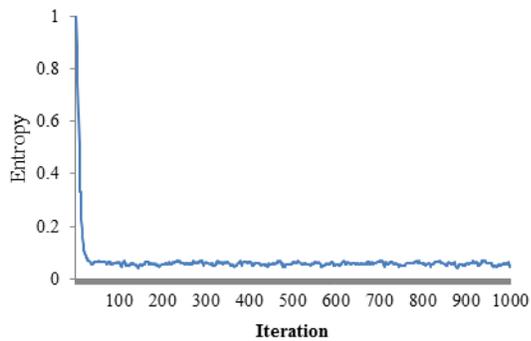

c) Dsjc500

According to the obtained results in figure 4, the downward entropy diagram shows effective information on convergence and is close to optimal value. Interestingly the proposed algorithm has fast convergence in experiments. In all instances and different performances full convergence with minimum number of nodes has been obtained with %100 success rate. Comparison of coverage in SA and ESA [17] methods for proposed algorithm is also shown in figure 5 as a bar chart diagram for P_hat3003.

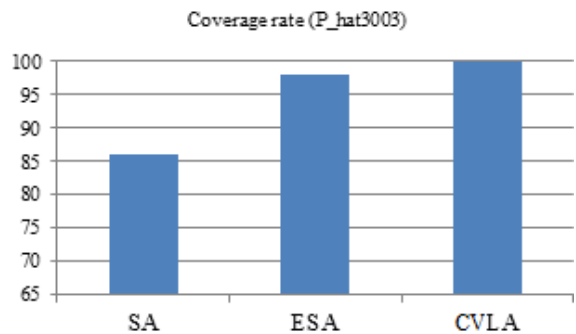

Fig. 5. Coverage Rate Bar Chart Comparison of Proposed Algorithm with Other Algorithms

Based on demonstrated results given in table 1, figures 4 and figure 5, the proposed algorithm has favorable performance in comparison to ESA, EWCC, EWLS and SA methods. Comparing proposed algorithm to other algorithms indicates that the number of covering nodes has came down which is a result of using less iteration than that of mentioned algorithms.

## V. CONCLUSION

In this paper, a new algorithm for minimum vertex cover problem was proposed using distributed learning automata. As mentioned above, most of the methods starting from a random set of optimal search whereas in proposed algorithm in each step define vertex set and improved the response using learning mechanism and good result have been obtained for minimum vertex cover problem. Due to exponential complexity of minimum vertex cover problem, the algorithm is not supposed to reach optimal solution in reasonable time, but the results show that proposed algorithm optimizes solutions than other well-known algorithms in minimum vertex cover problem.

## REFERENCES

[1] F.C. Gomes, C.N. Meneses, P. M. Pardalos, G. Valdisio, R. Viana, "Experimental analysis of approximation algorithms for the vertex cover and set covering problems," *Computers & Operations Research*, vol. 33, pp. 3520-3534, 2006.

[2] Y. Cheng Chen, W. Chih Peng, S.Yin Lee, "Efficient algorithms for influence maximization in social networks," *Knowledge and Information Systems*, vol. 33, pp. 577-601, 2012.

[3] S. Bouamama, C. Blum, A. Boukerram, "A population-based iterated greedy algorithm for the minimum weight vertex cover problem," *Applied Soft Computing*, vol. 12, pp. 1632-1639, 2012.

[4] C. Witt, "Analysis of an iterated local search algorithm for vertex cover in sparse random graphs," *Theoretical Computer Science*, vol. 425, pp. 117-125, 2012.

[5] J.Cardinal, M. Karpinski, "Approximating subdense instances of covering problems," *Electronic Notes in Discrete Mathematics*, vol. 37, pp. 297-302, 2011.

[6] M. Soleimani-Pouri, A. Rezvanian, M. R. Meybodi, "Solving maximum clique problem in stochastic graphs using learning automata," Proceedings of 2012 Fourth International Conference on Computational Aspects of Social Networks (CASoN), 2012, pp. 115-119.

[7] P. Horak, K. McAvaney, "On covering vertices of a graph by trees," *Discrete Mathematics*, vol. 308, pp. 4414-4418, 2008.

[8] S. Chiba, S. Fujita, "Covering vertices by a specified number of disjoint cycles, edges and isolated vertices," *Discrete Mathematics*, vol. 313, pp. 269-277, 2013.

[9] S. Richter, M. Helmert, C. Gretton, "A stochastic local search approach to vertex cover," Advances in Artificial Intelligence, vol. 4667, J. Hertzberg, Ed. Springer Berlin Heidelberg, 2007, pp. 412-426.

[10] H. Beigy, M. R. Meybodi, "Utilizing distributed learning automata to solve stochastic shortest path problem," *International Journal of Uncertainty, Fuzziness and Knowledge-based Systems*, vol. 14, no. 5, pp. 591-617, 2006.

[11] I. K. Evans, "Evolutionary algorithms for vertex cover, in Evolutionary Programming VII, vol. 1447, V. W. Porto, Ed. Springer Berlin Heidelberg, 1998, pp. 377-386.

[12] A. Rezvanian, M. R. Meybodi, "Tracking extrema in dynamic environments using a learning automata-based immune algorithm," in Grid and Distributed Computing, Control and Automation, vol. 121, T. Kim, Ed. Springer Berlin Heidelberg, 2010, pp. 216–225.

[13] M. Soleimani-Pouri, A. Rezvanian, M. R. Meybodi, "Finding a maximum clique using ant colony optimization and particle swarm optimization in social networks," Proceedings of the 2012 International Conference on Advances in Social Networks Analysis and Mining (ASONAM 2012), 2012, pp. 58-61.

[14] A. Rezvanian, M. R. Meybodi, "An adaptive mutation operator for artificial immune network using learning automata in dynamic environments" Proceedings of the 2010 Second World Congress on Nature and Biologically Inspired Computing (NaBIC 2010), 2010, pp. 479-483.

[15] R. Jovanovic, M. Tuba, "An ant colony optimization algorithm with improved pheromone correction strategy for the minimum weight vertex cover problem," *Applied Soft Computing*, vol. 11, pp 5360-5366, 2011.

[16] F. Amiri, N. Yazdani, H. Faili, and A. Rezvanian, "A novel community detection algorithm for privacy preservation in social networks," in Intelligent Informatics, vol. 18, A. Abraham, Ed. Springer Berlin Heidelberg, 2013, pp. 443–450.

[17] X. Xu, J. Ma, An efficient simulated annealing algorithm for the minimum vertex cover problem, *Neurocomputing*, vol. 69, pp. 913-916, 2006.

[18] E. Halperin, "Improved approximation algorithms for the vertex cover problem in graphs and hypergraphs," *SIAM Journal on Computing*, vol. 31, pp. 1608–1623, 2012.

[19] B. Escoffier, L. Gourves, J. Monnot, "Complexity and approximation results for the connected vertex cover problem in graphs and hypergraphs," *Journal of Discrete Algorithms*, vol. 8, pp. 36-49, 2010.

[20] S. Cai, K. Su, A .Sattar, "Local search with edge weighting and configuration checking heuristics for minimum vertex cover," Artificial Intelligence, Volume 175, pp.1672-1696, 2011.

[21] Z. Zhang, X. Gao, W. Wu, "Algorithms for connected set cover problem and fault-tolerant connected set cover problem," *Theoretical Computer Science*, vol. 410, pp. 812-817, 2009.

[22] R. Arakaki, and L. Lorena, "A constructive genetic algorithm for the maximal covering location problem, Proceedings of 4[th] Metaheuristics International Conference (MIC 2001), 2001, pp. 13-17.

[23] B. Masoumi, M. R. Meybodi, "speeding up learning automata based multi agent systems using the concepts of stigmergy and entropy," *Journal of Expert Systems with Applications*, vol. 38, no. 7, pp. 8105-8118, 2011.

[24] J. A. Torkestani, M. R. Meybodi, "Learning automata based algorithms for solving stochastic minimum spanning tree problem," *Journal of Applied Soft Computing*, vol. 11, no. 16, pp. 4064-4077, 2011.

[25] J. A. Torkestani, M. R. Meybodi, "Finding minimum weight connected dominating set in stochastic graph based on learning automata," Information Sciences, no. 200, pp.57-77, 2012.

[26] J. A. Torkestani, M. R. Meybodi, "A cellular learning automata based algorithm for solving the vertex coloring problem," *Expert Systems with Applications*, vol. 38, no. 8, pp. 9237-9247, 2011.